\documentclass[letterpaper, 10 pt, conference]{ieeeconf} 

\IEEEoverridecommandlockouts    
\usepackage[T1]{fontenc}
\usepackage[utf8]{inputenc}
\usepackage{graphicx}
\usepackage{amsmath}
\usepackage{amssymb}
\usepackage{float}
\usepackage{multirow}
\usepackage{array}
\usepackage[strict]{changepage}
\floatstyle{plaintop}
\restylefloat{table}
\usepackage{url}
\usepackage{hyperref}
\usepackage{microtype}
\usepackage{cite}
\usepackage{bm}

\usepackage[]{algorithm2e}

\usepackage{color}


\title{3D Human Pose Estimation on a Configurable Bed from a Pressure Image}

\author{Henry M. Clever*, Ariel Kapusta, Daehyung Park, Zackory Erickson, Yash Chitalia, Charles C. Kemp
\thanks{H. M. Clever, A. Kapusta, D. Park, Z. Erickson, Y. Chitalia and C. Kemp are with the Healthcare Robotics Lab, Institute for Robotics and Intelligent Machines, Georgia Institute of Technology,}
\thanks{*H. M. Clever is the corresponding author.
 {\tt\small henryclever@gatech.edu}}%
 }
 \renewcommand{\baselinestretch}{0.97}
\begin{document}
 \maketitle

\begin{abstract}


Robots have the potential to assist people in bed, such as in healthcare settings, yet bedding materials like sheets and blankets can make observation of the human body difficult for robots. A pressure-sensing mat on a bed can provide pressure images that are relatively insensitive to bedding materials. However, prior work on estimating human pose from pressure images has been restricted to 2D pose estimates and flat beds. In this work, we present two convolutional neural networks to estimate the 3D joint positions of a person in a configurable bed from a single pressure image. The first network directly outputs 3D joint positions, while the second outputs a kinematic model that includes estimated joint angles and limb lengths. We evaluated our networks on data from 17 human participants with two bed configurations: \textit{supine} and \textit{seated}. Our networks achieved a mean joint position error of 77~mm when tested with data from people outside the training set, outperforming several baselines. We also present a simple mechanical model that provides insight into ambiguity associated with limbs raised off of the pressure mat, and demonstrate that Monte Carlo dropout can be used to estimate pose confidence in these situations. Finally, we provide a demonstration in which a mobile manipulator uses our network's estimated kinematic model to reach a location on a person's body in spite of the person being seated in a bed and covered by a blanket. 

\end{abstract}

\section{Introduction} 
Various circumstances, such as illness, injury, or longterm disabilities can result in people receiving assistance in bed. Previous work has shown how robots can provide assistance with ADLs~\cite{hawkins2014,kapusta2016collaboration, grice2016assistive}, but providing assistance to a person in bed can be challenging. Estimating the pose of a person's body could enable robots to provide better assistance. Typical methods of body pose estimation use line-of-sight sensors, such as RGB cameras, which can have difficulties when the body is occluded by blankets, loose clothing, medical equipment, over-bed trays and other common items in healthcare settings, such as hospitals. A pressure-sensing mat on the bed can allow for estimation of the body's pose in a manner that is less sensitive to bedding materials and surrounding objects \cite{kapusta2016collaboration,harada2001pressure, grimm2012markerless,liu2014bodypart}. However, prior work with pressure images has not addressed a number of concerns key to the success of robot assistance in bed, namely (1) pose estimation in 3D for either flat or non-flat beds and (2) appropriately dealing with uncertainty when the pose estimate may be inaccurate. 

In this work, we present a method for estimating the 3D joint positions in real time with a measure of confidence in each estimated position for a person in a configurable bed using a pressure-sensing mat. 
We provide evidence that our method works for some challenging scenarios, such as when the bed and human are configured in a seated posture, and when limbs are raised off of the pressure sensing mat. Further, we release a motion capture labeled dataset of over 28,000 pressure images across 17 human participants, in addition to our open-source code.

\begin{figure}[t!]
\centering
\includegraphics[width=\columnwidth]{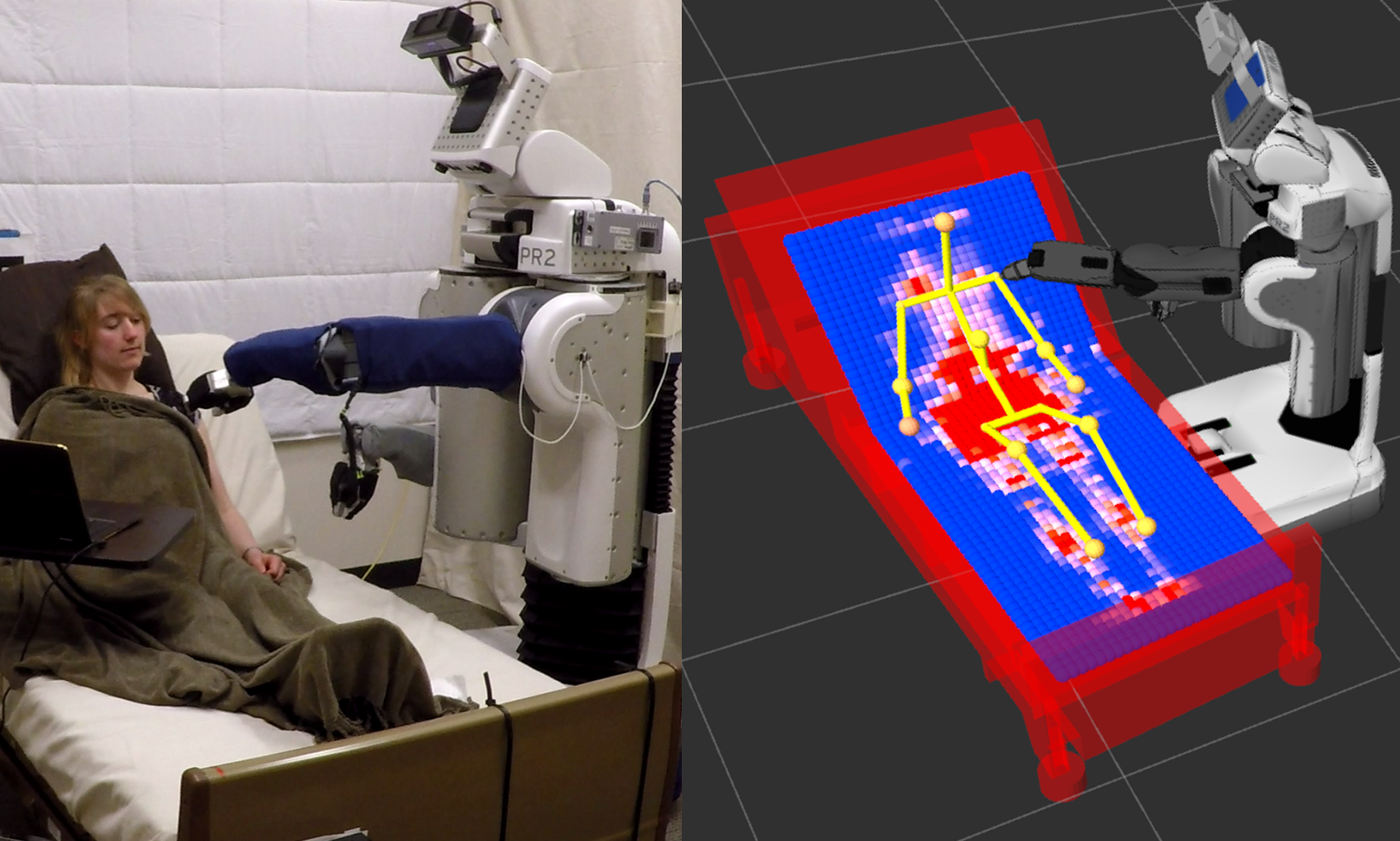}
\caption{We demonstrate how an assistive robot could use our 3D human pose estimation method. A PR2 robot uses our method's body pose estimation to reach to a person's shoulder.}
\vspace{-3mm}
\label{fig:demo_shoulder}
\end{figure}

In prior work, we used a pressure sensing mat on a configurable bed to estimate the location of a person's head for positioning a rigid geometric model of a human body~\cite{kapusta2016collaboration}. In this paper, we propose and compare two convolutional neural network (ConvNet) architectures to learn a mapping from a pressure image and bed configuration to 3D human joint positions: the head, neck, shoulders, elbows, wrists, chest, hips, knees, and ankles. 
The first method directly regresses to the 3D ground truth labels (x, y, z). The second method embeds a 17-joint kinematics skeleton model of the human to the last layer of the ConvNet to enforce constraints on bone lengths and joint angles. This provides a more complete pose representation with additional unlabeled joints and latent space joint angle estimates. We introduce a new architecture that adjusts the skeleton model for differently sized people, while providing comparison to a constant link length skeleton model. We train the kinematics ConvNets end-to-end, backpropagating from 3D joint Euclidean error through the kinematics model. We compare our ConvNet methods to baseline data-driven algorithms, including ridge regression and k-nearest neighbors.

Our configurable bed, introduced previously as Autobed~\cite{kapusta2016collaboration,grice2016autobed}, features adjustable height, leg rest angle, and head rest angle. Autobed can sense its own state, sense the pressure distribution of the person on the bed, and communicate with other devices. In this work, we estimate the human joint positions in two Autobed configurations by adjusting the bed's head rest: supine ($0^{\circ}$ flat) and seated ($60^{\circ}$ incline). 

Lack of contact by limbs or other body parts presents a challenging issue for the pressure image modality. In this work, we consider common poses where this issue arises, such as an arm raised in the air resembling a double inverted pendulum. Among other poses, this case demonstrates where the pressure image can be similar for different configurations of the arm. In such a case, the pressure data may be insufficient to confidently estimate the pose of the arm.
An estimate of confidence or model uncertainty can be valuable, for example to allow an assistive robot to reject low-confidence estimates by removing them from a list of potential goals in task plan execution. 
To estimate model uncertainty, we use Monte Carlo dropout, a method proposed by Gal and Ghahmarani \cite{gal2016dropout}. With this method, we perform a number of stochastic forward passes through the ConvNet during test time and compute the joint position and joint position confidence from the moments of the output distribution.

\section{Related work}\label{sec:lit_review}

Markerless human pose estimation is a challenging problem complicated by environment factors, the human pose configurations of interest, and data type. 
Relatively few researchers have used pressure images for human pose estimation in bed \cite{harada2001pressure,grimm2012markerless,liu2014bodypart}, while many used cameras in myriad environments and poses\cite{gong2012human,sarafianos20163dhuman,okada2008relevant,ionescu2014human,agarwal2006recovering,toshev2014deep,tompson2014joint,pavlakos2017coarse,zhou2016deep,wei2016convolutional,zhou2016sparseness,bogo2016keep,li20143dhuman}. Researchers have increasingly explored data-driven methods such as ConvNets, from model-free networks to inclusion of models in various architectures. Here we discuss research with pressure images, data-driven methods, and measuring network uncertainty.

\subsection{Pressure-Image-based Work}
Prior 
pressure-image-based pose estimation work has fit 2D kinematic models to pressure image features. In a series of papers including \cite{harada2001pressure}, Harada et al. used a kinematic model to create a synthetic database for comparison with ground truth pressure images. Grimm et al.\cite{grimm2012markerless} identified human orientation and pose using a prior skeleton model. Similar to our motivation and findings, they used a pressure mat to compensate for bedding occlusion and observed higher error for lighter joints (e.g. the elbow), which had a relatively low pressure. Liu et al. \cite{liu2014bodypart} generated a pictorial structures model to localize body parts on a flat bed.

A few researchers have also looked at human posture classification from pressure images \cite{grimm2012markerless,farshbaf2013detecting,ostadabbas2014bed}. Posture classification is a different problem from 3D body pose estimation, but it can be used in a complementary way, e.g. by providing a prior on the model used for pose estimation.

\subsection{Data-driven Human Pose Estimation}

Like the pressure-image-based research, we use a human body model, but take a data-driven approach more common in vision-based work. 
While infrared and depth images have seen recent attention in human pose estimation \cite{gong2012human}, a large body of vision-based work uses monocular RGB cameras \cite{sarafianos20163dhuman}. Our method builds upon research with monocular RGB image input; we note the following similarities and differences:
\begin{itemize}

\item \textit{Single image.} Both monocular RGB-image-based work and our pressure-image-based work has a single input array. 
\item \textit{Under-constrained.} For 3D human pose, both monocular RGB images and single pressure images are under-constrained.
\item \textit{Data content.} The data encoding is fundamentally different. For example, in the context of pose estimation, light intensity in an RGB image is highly disconnected from pressure intensity. 
\item \textit{Dimensionality.} RGB images typically have more features and a higher resolution than pressure images.
\item \textit{Warped spatial representation.} Calibrating RGB cameras to alleviate distortion is straightforward. In contrast, it is challenging to determine the configuration of a cloth pressure sensing mat from its pressure image in the case of folds or bends.

\end{itemize}


While the differences may limit the transferability of methods across data types, we find that some data-driven methods previously used for vision modalities are applicable to pressure images. Researchers have performed 3D human pose estimation with monocular RGB images using standard machine learning algorithms such as ridge regressors \cite{okada2008relevant,ionescu2014human,agarwal2006recovering}. In particular, Okada and Soatto \cite{okada2008relevant} use kernel ridge regression (KRR) as well as linear ridge regression (LRR). Further, Ionescu et al. \cite{ionescu2014human} compare K-Nearest Neighbors (KNN) and ridge regression on the Human3.6M dataset. We use these classical approaches to provide a baseline comparison for our proposed method.

Recently, with the advent of high quality, labeled synchronous datasets such as Human3.6M \cite{ionescu2014human}, many researchers have explored deep learning methods such as end-to-end training of ConvNets \cite{toshev2014deep,tompson2014joint,pavlakos2017coarse,zhou2016deep}.
Two common ConvNet approaches include direct regression to joint labels \cite{toshev2014deep, li20143dhuman} and regression to discretized confidence maps \cite{tompson2014joint, wei2016convolutional, pavlakos2017coarse}. Within 3D human pose estimation research, confidence map approaches include Pavlakos et al. \cite{pavlakos2017coarse}, who train a ConvNet end-to-end on a 3D confidence voxel space, and Zhou, Zhu et al. \cite{zhou2016sparseness} and Bogo et al. \cite{bogo2016keep} who fit a 3D model to 2D confidence maps. However, the high dimensional output space of confidence maps can make real-time pose estimation difficult. Li and Chan \cite{li20143dhuman} used rapid direct regression to 3D Cartesian joint positions; we implement a similar architecture because real-time estimation is important to our planned application. Zhou, Sun et al. \cite{zhou2016deep} take a hybrid approach, by training a ConvNet end-to-end and enforcing anthropomorphic constraints with an embedded human skeleton kinematics model with constant link lengths. We implement a method of this form for comparison and introduce a new architecture with variable skeleton link lengths to allow the model to adapt to differently sized people.

\begin{figure*}
\centering
\vspace{4mm}
\includegraphics[width=\textwidth]{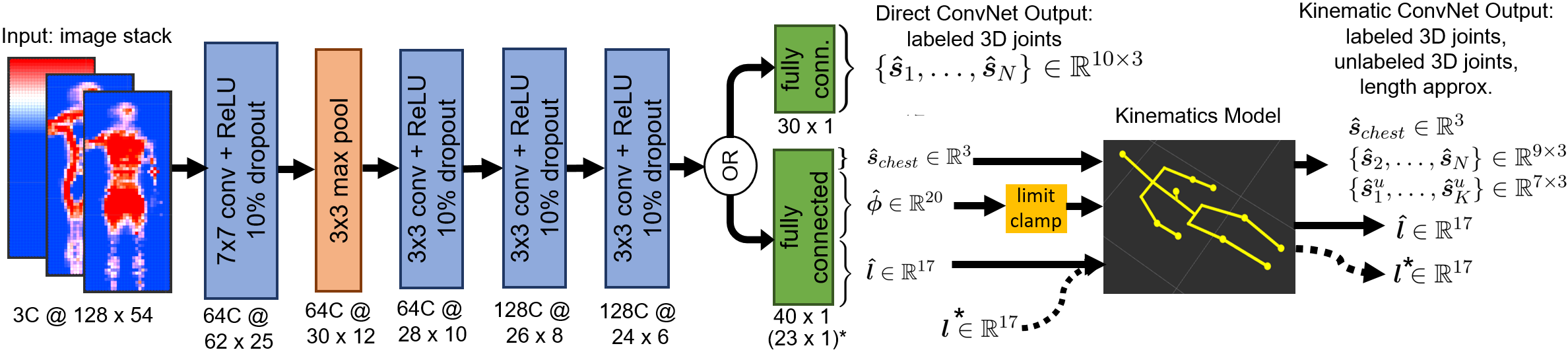}
\caption{Two ConvNet architectures. The \textit{Direct ConvNet} directly regresses to ten motion capture labeled global joint positions. The \textit{Kinematic ConvNet} embeds a kinematics skeleton model into the last fully connected network layer, parameterized in the latent space by a root joint global position, joint angles, and skeleton link lengths. The \textit{Kinematic ConvNet} also outputs unlabeled joint position estimates. We explore architectures with both variable link length (shown) and constant link length (dashed arrows, defined by $*$.)}
\vspace{-4mm}
\label{fig:convnet_kinematics}
\end{figure*}

\subsection{Monte Carlo Dropout in ConvNets}
We use Monte Carlo dropout to measure network uncertainty, a method introduced by Gal and Ghahramani \cite{gal2016dropout}. Monte Carlo dropout has been applied to measure uncertainty for camera relocalization \cite{kendall2016modelling} and semantic segmentation \cite{kampffmeyer2016semantic}. We use this method to estimate pose and provide a measure of confidence.

\section{Method}\label{sec:method}
Our ConvNets learn a function $f(\mathcal{P}, \theta_B)$ that estimates pose parameters of a person lying in a robotic bed, given a specified bed configuration $\theta_B$ and a 2D pressure image $\mathcal{P}$ from a pressure mat.

\subsection{ConvNet Architecture}\label{ssec:convnet}
We explore two ConvNet architectures shown in Fig.~\ref{fig:convnet_kinematics}. Our network includes four 2D convolutional layers with 64 output channels for the first two layers and 128 channels for the last two layers. The layers mostly have $3 \times 3$ filters, with a ReLU activation and a dropout of 10\% applied after each layer. We apply max pooling, and the network ends with a linear fully connected layer. 

To estimate a person's joint pose, we construct an input tensor for the ConvNet comprised of three channels, i.e. $\{\mathcal{P}, E, B\}\in\mathbb{R}^{128 \times 54 \times 3}$. Raw data from the pressure sensing mat is recorded as a $(64 \times 27)$-dimensional image, which we upsample by a factor of two, i.e. $\mathcal{P}\in\mathbb{R}^{128 \times 54}$. We use first order interpolation for upsampling. In addition to a pressure image, we also provide the ConvNet with an edge detection channel, $E\in\mathbb{R}^{128 \times 54}$, which is computed as a Sobel filter over both the horizontal and vertical directions of the upsampled image. Empirically, we found this edge detection input channel improved pose estimation performance. In order to estimate human pose at different configurations of the bed (e.g. sitting versus lying down), we compute a third input channel, $B\in\mathbb{R}^{128\times 54}$, which depicts the bed configuration. Specifically, each element in the matrix $B$ depicts the vertical height of the corresponding taxel on the pressure mat. When the bed frame is flat, i.e. $\theta_B=0$, then $B$ is simply the zero matrix.

\subsection{Direct Joint Regression}
The first proposed ConvNet architecture outputs an estimate of the motion capture labeled global 3D joint positions $\{\bm{\hat{s}}_1, \ldots, \bm{\hat{s}}_N\}$, 
where each $\bm{\hat{s}}_j \in \mathbb{R}^{3}$ represents a 3D position estimate for joint $j$. This direct ConvNet regresses directly to 3D ground truth label positions in the last fully connected layer of the network. We compute the loss from the absolute value of Euclidean error on each joint:
\begin{equation}
\text{Loss}_{\text{direct}} = \sum_{j=1}^{N}||\boldsymbol{s}_j - \boldsymbol{\hat{s}}_j||
\end{equation}\label{eq:lossdir}

\begin{figure}
\centering
\includegraphics[width=\columnwidth]{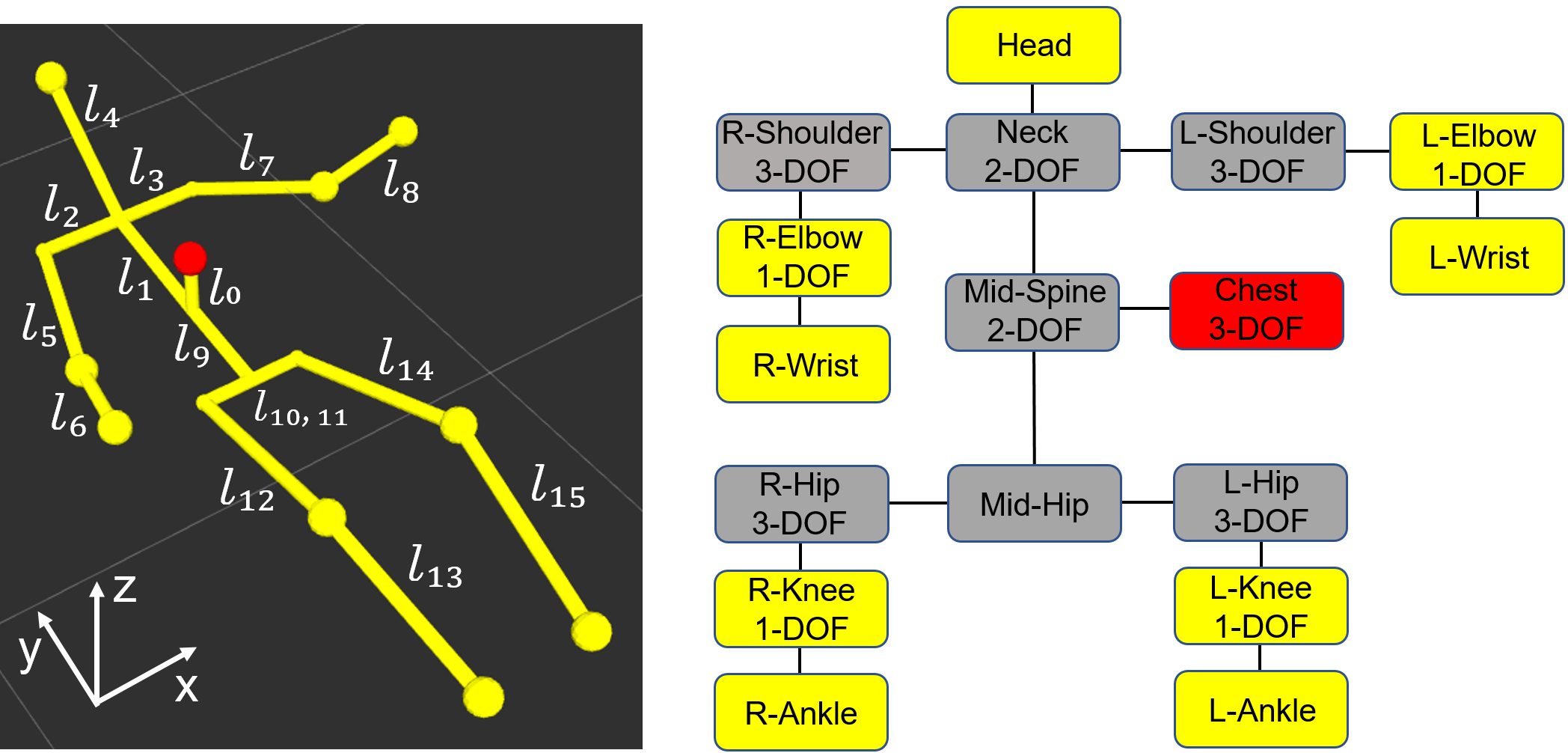}
\caption{Kinematics Model Parameters. The root joint $\boldsymbol{s}_{chest}$ is specified in red. Yellow boxes represent the remaining motion capture labeled joints $\{\bm{\hat{s}}_2, \ldots, \bm{\hat{s}}_N\}$. Grey boxes represent unlabeled joints $\{\bm{\hat{s}}^u_1, \ldots, \bm{\hat{s}}^u_K\}$, where the kinematic model adjusts to an approximate fit.}
\vspace{-3mm}
\label{fig:skeleton_parameters}
\end{figure}

\subsection{Deep Kinematic Embedding}\label{ssec:kincnn}
We embed a human skeleton kinematics model into the last fully connected network layer to enforce geometric and anthropomorphic constraints. This creates an extra network layer where labeled joint estimates $\{\bm{\hat{s}}_1, \ldots, \bm{\hat{s}}_N\}$ and unlabeled joint estimates $\{\bm{\hat{s}}^u_1, \ldots, \bm{\hat{s}}^u_K\}$ are solved through forward kinematics equations depending on root joint position $\bm{\hat{s}}_{1}$, latent space joint angles $\bm{\hat{\phi}}$ and an estimate of skeleton link length approximations $\bm{\hat{l}}$. We use $\bm{\hat{s}}_{1}=\bm{\hat{s}}_{chest}$ as the root joint. We incorporate skeleton link lengths $\bm{l}$ into the loss function by pre-computing an approximation to the ground truth. While the kinematics functions are relative to a root joint, we learn root joint global position $\boldsymbol{s}_{chest}$ to put our output in global space. We compute a weighted loss from the absolute value of Euclidean error on each joint and error on each link length:
\begin{equation}
\text{Loss}_{\text{kin.}} = ||\boldsymbol{s}_{1} -\boldsymbol{\hat{s}}_{1}||+ \alpha\sum_{j=2}^{N}||\boldsymbol{s}_j - \boldsymbol{\hat{s}}_j|| +\beta|\boldsymbol{l} - \boldsymbol{\hat{l}}|\label{eq:lossconv}
\end{equation}
where $\alpha$ and $\beta$ are weighting factors. We compare two variants of this loss function: The \textit{variable link length} ConvNet is as described, while for the \textit{constant link length} ConvNet we set $\beta = 0$ and use a constant $\boldsymbol{l}^*$ input to the kinematics model. We compute $\boldsymbol{l}^*$ as the average of the approximations $\bm{l}$.

\subsubsection{Human Kinematics Model}\label{sssec:kinmodel}
We represent the human body with a model similar to that used in other work \cite{ionescu2014human,zhou2016deep,li20143dhuman,mehta2017vnect}, with 17 joints to cover major links down to the wrists and ankles. We ignore minor links and joints. 
To train the networks that have link lengths as an output, we require ground truths for comparison. Some ground truth link lengths may be calculated directly from the dataset's labels, for example when motion capture gives the location of both ends of the link. We approximate the link lengths for links that are under-constrained in the dataset for the skeleton model. The link lengths are an output of our network as a $\boldsymbol{l} \in \mathbb{R}^{17}$ vector. We ignore unlabeled joints in the loss function.

The mid-spine is found by a vertical offset from the chest marker to compensate for the distance between marker placement atop the chest and the modeled bending point of the spine. We do not make offset corrections with other joints; these are more challenging than the chest and the effects are less noticeable. 
 
\subsubsection{Angular Latent Space}\label{sssec:netimpl}
We define 20 angular degrees of freedom consisting of 3-DOF shoulders and hips, 1-DOF elbow and knee joints, a 2-DOF spine joint, and a 2-DOF neck joint. Fig \ref{fig:skeleton_parameters} (b) shows this parameterization corresponding to labeled and unlabeled joints. For the spine of the model to better match the spine of a person seated in bed, we used two revolute joints about the $x$-axis. To account for head movement, we placed a neck joint at the midpoint of the shoulders with pitch and yaw rotation. We use PyTorch \cite{paszke2017automatic}, a deep learning library with tensor algebra and automatic differentiation. We manually encoded the forward kinematics for the skeleton kinematic model.
The network uses stochastic gradient descent during backpropagation to find inverse kinematics (IK) solutions.

\begin{figure}[t!]
\centering
\vspace{4mm}
\includegraphics[width=\columnwidth]{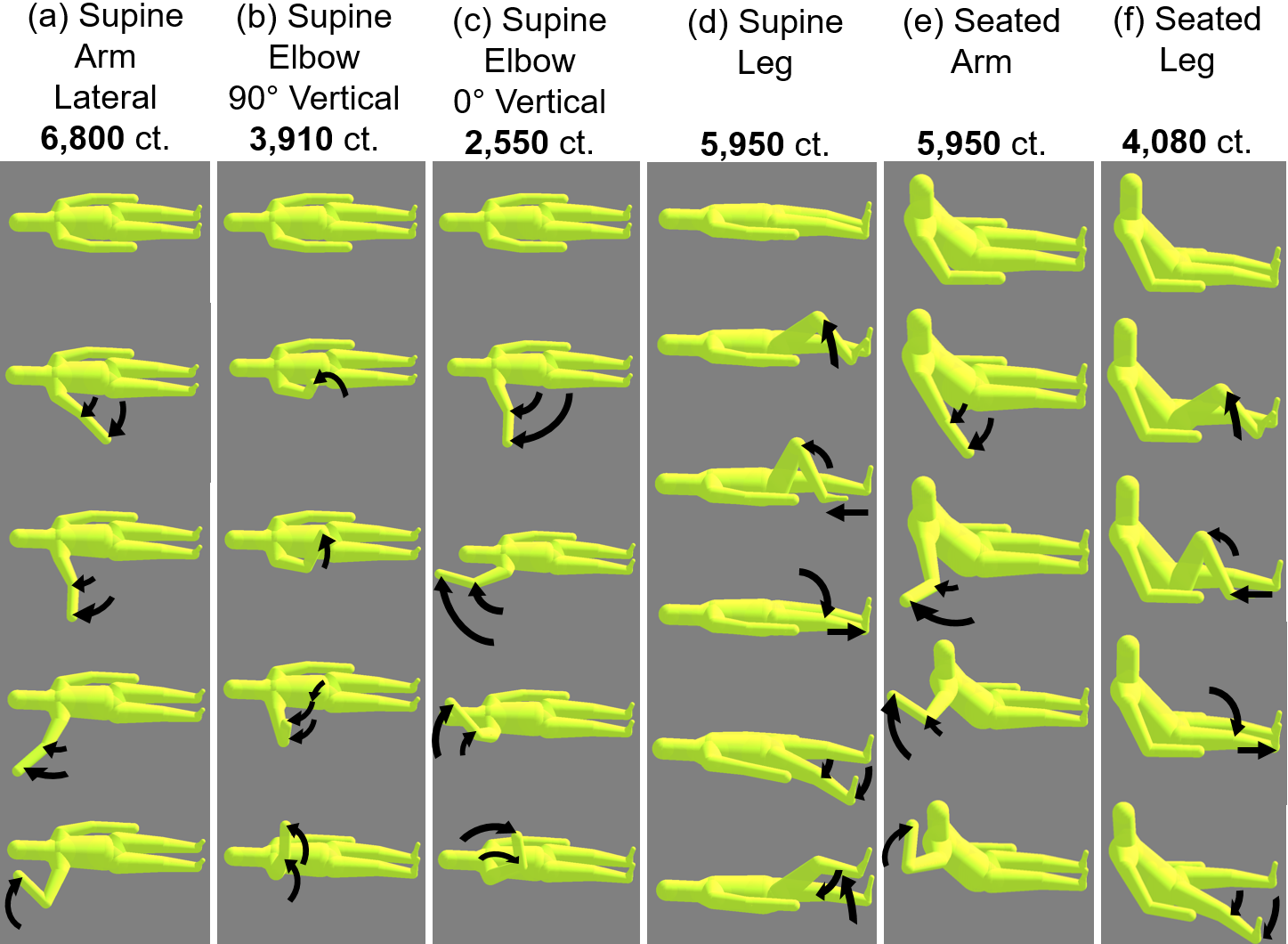}
\caption{Range of paths traversed by participants during training. Equivalent paths were traversed by the left limbs. Count represents both left and right data across 17 participants.}
\vspace{-3mm}
\label{fig:training_poses_supine}
\end{figure}

\subsection{Pressure Image Ambiguity}
\begin{figure}[t!]
\centering
\vspace{4mm}
\includegraphics[width=\columnwidth]{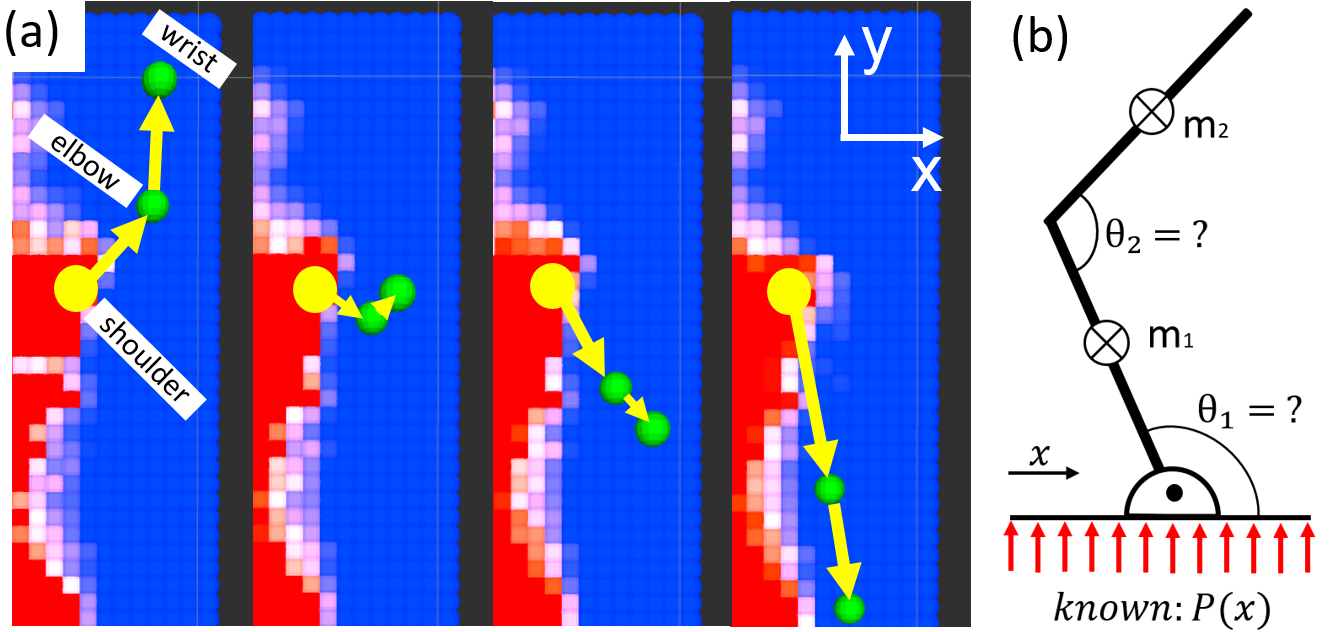}
\caption{(a) Crossover of the shoulder from the \textit{supine elbow 0 vertical} traversal, both joints envisioned as 2-DOF inverted pendulum. Green dots represent left elbow and wrist ground truth markers projected in 2D, yellow dot and arrows indicate approximate shoulder and limb positions. (b) Model of 2-DOF inverted pendulum showing static indeterminacy. Known pressure distribution $P(x)$ is insufficient to solve $\theta_1, \theta_2$.}
\vspace{-3mm}
\label{fig:ambiguity_illustration}
\end{figure}

Raising a limb off of a bed with pressure sensors can lead to a loss of information as the sensors can only sense pressure during contact. A similar loss of information can be seen when the limbs extend off the edge of the bed. Consider the movement shown in Fig.~\ref{fig:training_poses_supine}~(c) and an example of the pressure images associated with such a movement in Fig.~\ref{fig:ambiguity_illustration} (a). Here, the pressure images appear nearly identical while the elbow and wrist positions change substantially. To better understand what is physically causing this phenomena, we can model the arm as a double inverted pendulum, shown in Fig. \ref{fig:ambiguity_illustration} (b). Here, the pendulum angles $\theta_1$ and $\theta_2$ are statically indeterminate given an underlying pressure distribution, until part of the arm touches the sensors.

Another challenge is pressure sensor resolution. Depending on the type of sensor, saturation can occur. A pressure image with higher spatial resolution, accuracy, and pressure range may result in less ambiguity. We note that a model of body shape (e.g. 3D limb capsules) might provide information that helps to resolve the ambiguity.

\subsection{Uncertainty: Monte Carlo Dropout}
To estimate both joint position and model uncertainty simultaneously, we apply Monte Carlo dropout from Gal and Ghahramani \cite{gal2016dropout}. Monte Carlo dropout is the process of performing $V$ forward passes through the network with dropout enabled. This results in $V$ output vectors which may differ slightly due to the stochastic dropout of data during each forward pass. We can compute an estimated output as the average of all $V$ outputs, corresponding to the first moment of the predictive distribution within the network. Similarly, the model's uncertainty corresponds to the second moment of the distribution, which we can compute as the variance of all $V$ forward passes. 

\section{Evaluation}\label{sec:evaluation}

We recorded a motion capture labeled dataset with over 28,000 pressure images from 17 different human participants.\footnotemark
\footnotetext{Dataset: ftp://ftp-hrl.bme.gatech.edu/pressure\char`_mat\char`_pose\char`_data}
We conducted this study with approval from the Georgia Institute of Technology Institutional Review Board (IRB), and obtained informed consent from all participants. We recruited 11 male and 6 female participants aged 19-32, who ranged 1.57-1.83 m in height and 45-94 kg in weight. We fitted participants with motion capture markers at the wrists, elbows, knees, ankles, head, and chest. We used a commercially available $64\times 27$ pressure mat from Boditrak sampled at 7~Hz. We asked each participant to move their limbs in 6 patterns, 4 while supine and 2 while seated, to represent some common poses in a configurable bed. The movement paths are shown in Fig.~\ref{fig:training_poses_supine}. We instructed participants to keep their torso static during limb movements.

We trained six data-driven models: three baseline supervised learning algorithms and the three proposed ConvNet architectures.\footnotemark
\footnotetext{Code release: https://github.com/gt-ros-pkg/hrl-assistive/tree/indigo-devel/hrl\char`_pose\char`_estimation} We designed the network using 7 participants (5M, 2F); we performed leave-one-participant-out cross validation using the remaining 10 participants (6M, 4F).

\subsection{Data Augmentation}\label{ssec:augmentation}

At each training epoch for the ConvNets, we selected images such that each participant would be equally represented in both the training and test sets. We augmented the original dataset in the following ways to increase training data diversity: 

\begin{itemize}

\item \textit{Flipping.} Flipped across the longitudinal axis with probability $P = 0.5.$
\item \textit{Shifting.} Shifted by an additive factor $sh \sim \mathcal{N}(\mu = 0 cm,\sigma = 2.86 cm)$.
\item \textit{Scaling.} Scaled by a multiplicative factor $sc \sim \mathcal{N}(\mu = 1,\sigma = 0.06)$.
\item \textit{Noise.} Added taxel-by-taxel noise to images by an additive factor $\mathcal{N}(\mu = 0,\sigma = 1)$. Clipped the noise at min pressure (0) and at the saturated pressure value (100).
\end{itemize}
We chose these to improve the network's ability to generalize to new people and positions of the person in bed. We did not shift the seated data longitudinally or scale it because of the warped spatial representation.

\subsection{Baseline Comparisons}\label{ssec:baselines}
We implemented three baseline methods to compare our ConvNets against: K-nearest neighbors (KNN), Linear Ridge Regression (LRR), and Kernel Ridge Regression (KRR). For all baseline methods, we used histogram of oriented gradients (HOG) features~\cite{dalal2005histograms} on $2\times$ upsampled pressure images. We applied flipping, shifting, and noise augmentation methods. We did not use scaling, as it worsened performance.

\subsubsection{K-Nearest Neighbors}\label{sssec:knn}
We implemented a K-nearest neighbors (KNN) regression baseline using Euclidean distance on the HOG features to select neighbors, as \cite{ionescu2014human} did. We selected $k = 10$ for improved performance. 

\subsubsection{Ridge Regression}\label{ridge}
We implemented two ridge-regression-based baselines. Related work using these methods are described in Section~\ref{sec:lit_review}.
We trained linear ridge regression (LRR) models with a regularization factor of $\alpha = 0.7$. We also train Kernel Ridge Regression (KRR) models with a radial basis function (RBF) kernel and $\alpha = 0.4$. We manually selected these values of $\alpha$ for both LRR and KRR. We also tried linear and polynomial kernels for KRR, but found the RBF kernel produced better results in our dataset.

\subsection{Implementation Details of Proposed ConvNets}\label{convneteval}
During testing, we estimated the joint positions with $V = 25$ forward passes on the trained network with Monte Carlo dropout for each test image. For each joint, we report the mean of the forward passes as the estimated joint position. We use PyTorch and ADAM from \cite{kingma2014adam} for gradient descent.

\subsubsection{Pre-trained ConvNet}
We created a pre-trained ConvNet that we use to initialize both Kinematic ConvNets, with regressed and constant link length. The pre-trained network used the kinematically embedded ConvNet and the loss function in Eq.~\ref{eq:lossconv}, with $\alpha = 0.5$ and $\beta = 0.5$. This network was trained for 10 epochs on the 7 network-design participants with a learning rate of 0.00002 and weight decay of 0.0005.

\subsubsection{Direct ConvNet}
We trained the network for 300 epochs directly on motion capture ground truth, using the sum of Euclidean error as the loss function. We used a learning rate of 0.00002 and a weight decay of 0.0005.

\subsubsection{Kinematic ConvNet, Constant Link Length}
We trained the network through the kinematically embedded ConvNet, used the loss function in Eq.~\ref{eq:lossconv}, with $\alpha = 0.5$ and $\beta = 0$. This value for $\beta$ means the network would not regress to link length, leaving it constant. We initialized the network with the pre-trained ConvNet, but we separately initialized each link length as the average across all images in the fold's training set for each fold of cross validation. 

\subsubsection{Kinematic ConvNet, Regress Link Length}
We trained the network through the kinematically embedded ConvNet, used the loss function in Eq.~\ref{eq:lossconv}, with $\alpha = 0.5$ and $\beta = 0.5$. Joint Cartesian positions and link lengths in the ground truth are represented on the same scale. We initialized with the pre-trained ConvNet.

\subsection{Measure of Uncertainty}\label{ssec:result_obj_all_recog}
Here we show an example where ambiguous pressure mat data has a high model uncertainty. We compare two leg abduction movements from the \textit{supine leg} motion, shown in the bottom two columns of Fig. \ref{fig:training_poses_supine} (d). We sample 100 images per participant: half feature leg abduction contacting the pressure mat, and half with elevated leg abduction. For each pose, we use $V=25$ stochastic forward passes and compute the standard deviation of the Euclidean distance from the mean for abducting joints, including knees and feet. We compare this metric between elevated and in-contact motions.

\begin{figure*}
\centering
\vspace{4mm}
\includegraphics[width=\textwidth]{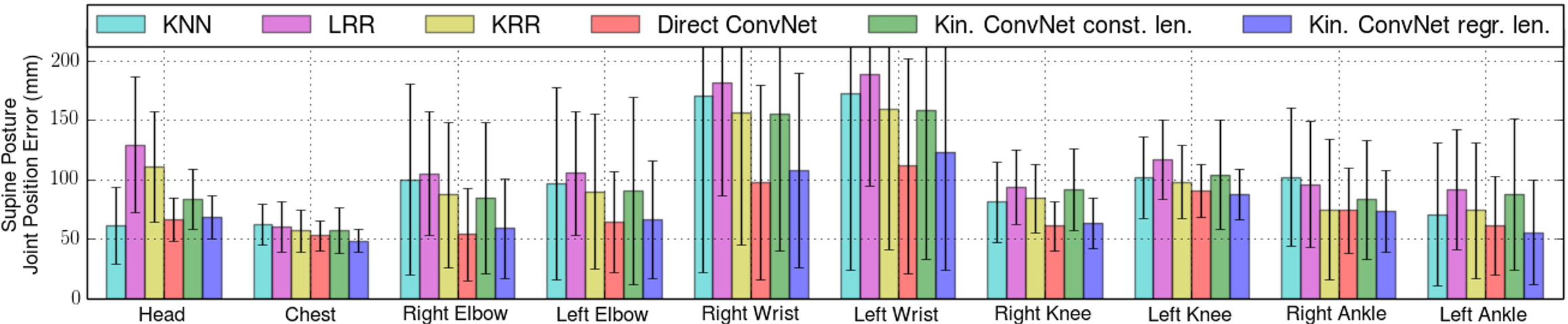}
\includegraphics[width=\textwidth]{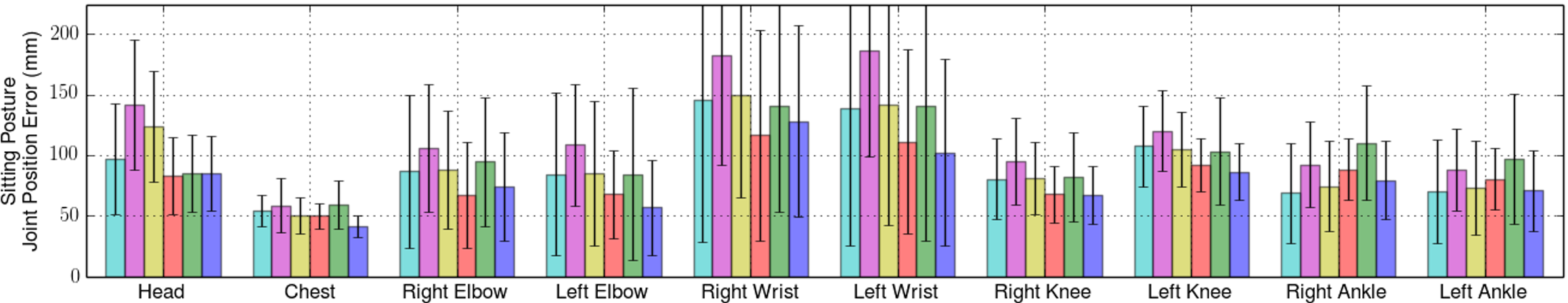}
\caption{Per-joint error mean and standard deviation for leave-one-participant-out cross validation over 10 participants for a sitting and a supine posture in bed. Lower is better. Our methods outperformed baseline methods.
}
\vspace{-2mm}
\label{fig:supine_error}
\end{figure*}

\section{Results}\label{sec:expt}
\begin{figure}
\centering
\includegraphics[width=2.5in]{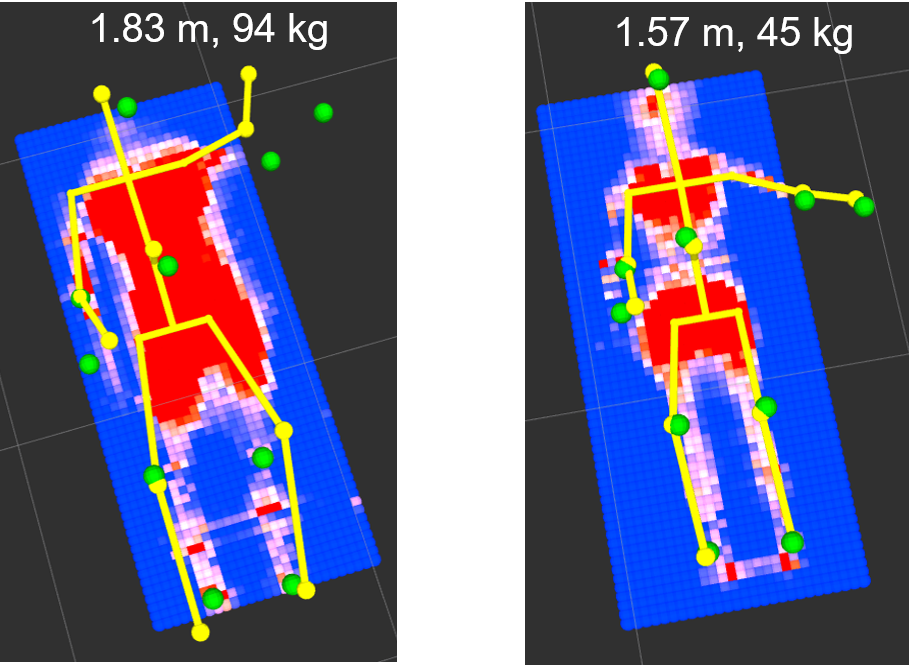}
\caption{Comparison of the heaviest and tallest participant with the lightest and shortest participant. Our kinematics ConvNet with link length regression appears to adjust for both sizes. }
\vspace{-3mm}
\label{fig:pose_shots}
\end{figure}

In Table~\ref{table:error} , we present the mean per joint position error (MPJPE), a metric from literature to represent overall accuracy~\cite{li20143dhuman, ionescu2014human}. Fig. \ref{fig:supine_error} shows the per-joint position error across all trained models, separated into supine and seated postures. 
The error for the direct ConvNet and the kinematic ConvNet with length regression is significantly lower than the other methods. The results of knees and legs show that more distal limbs on the kinematic chain do not necessarily result in higher error. The wrists are both distal and light, and have higher error than the other joints. Fig. \ref{fig:pose_shots} shows the kinematics ConvNet with length regression adjusting for humans of different sizes and in different poses. 
Furthermore, we can perform a pose estimate with uncertainty using $V=25$ stochastic forward passes in less than a half second.

\begin{table}[t!]
\begin{center}
\footnotesize
\caption{Mean Per Joint Position Error. }\label{tbl:seated}
\renewcommand{\arraystretch}{1.1}
\vspace{1mm}

\begin{tabular} {c|c|c|c}
 & MPJPE & MPJPE & MPJPE \\
 & Supine& Seated & Overall\\
Method & (mm) & (mm) & (mm)\\
\hline
\hline
K-Nearest Neighbors & 102.01 & 93.42 &99.06\\
\hline
Linear Ridge Regression &117.01&114.76&116.24\\
\hline
Kernel Ridge Regression &99.25 &97.10&98.51\\
\hline
Direct ConvNet & 73.49 & 82.44 & \textbf{76.56}\\
\hline
Kinematic ConvNet, avg. $\boldsymbol{l}$ &99.74 & 99.70 & 99.72\\
\hline
Kinematic ConvNet, regr. $\boldsymbol{l}$ & 75.43 & 79.19 & \textbf{76.72}\\
\cline{1-4}
\end{tabular}
\label{table:error}
\end{center}
\end{table}

\subsection{Measure of Uncertainty}\label{ssec:result_obj_all_recog}
\begin{figure*}
\centering
\vspace{4mm}
\includegraphics[width=\textwidth]{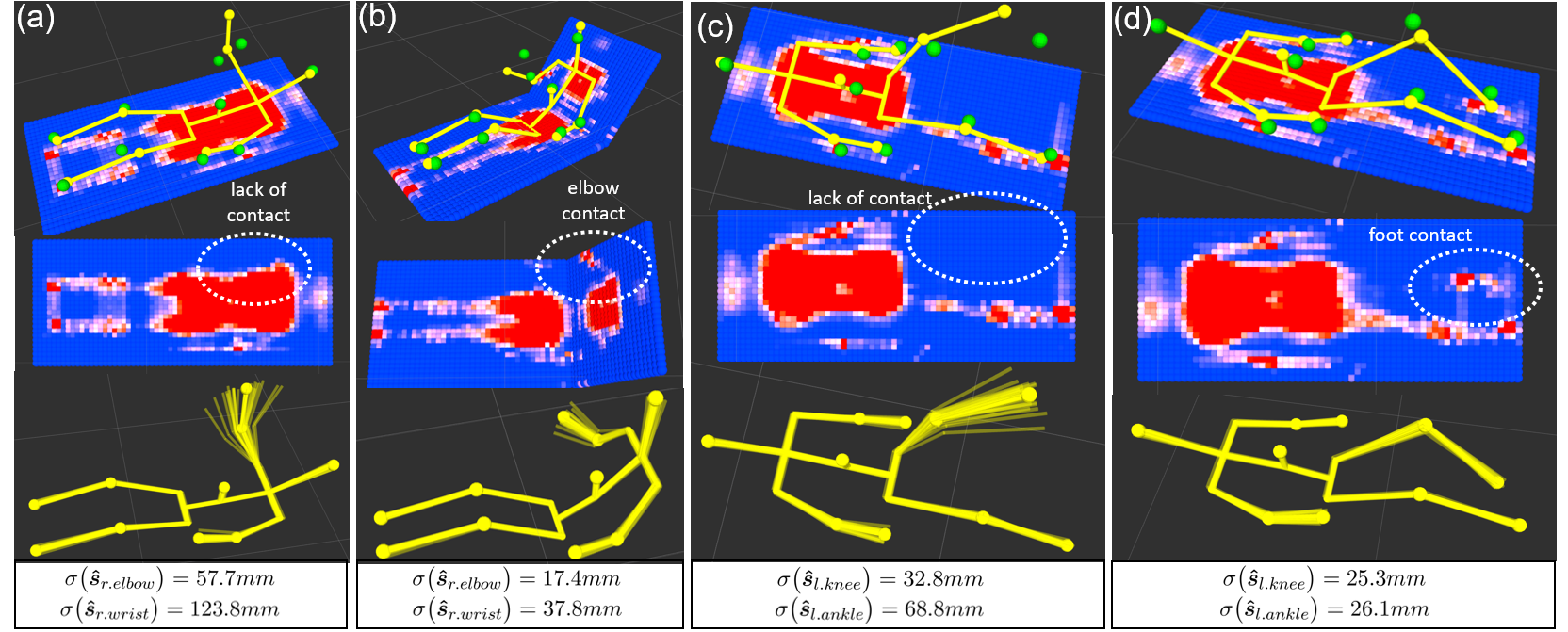}
\caption{Illustration of mean and standard deviation of kinematic ConvNet ($\boldsymbol{l}$ regression) output. $V=25$ forward passes with Monte Carlo dropout shown with thin translucent skeleton links. Spheres indicate joint position estimates.
(a) Right arm thrust into the air. (b) Right forearm extending off the side of the pressure mat. (c) Left leg extended in the air and abducted. (d) Left knee extended in the air, left foot contacting pressure mat. Note: (c) and (d) show the same participant, others are different.}
\vspace{-4mm}
\label{fig:confidence}
\end{figure*}

We performed a t-test to compare uncertainty in elevated leg abduction and in-contact leg abduction.  We compared each knee and ankle separately. We found that the standard deviation of the Euclidean distance from the mean of $V=25$ forward passes with Monte Carlo dropout is significantly higher for joints in the elevated position. Further, we note that variance in the latent angle parameters $\boldsymbol{\theta}$ compounds through the kinematic model, causing more distal joints in the kinematic chain to have higher uncertainty. This phenomena is further described in Fig. \ref{fig:confidence}, which shows limbs removed from the mat that have a high variance. 

\section{Discussion and Limitations}\label{sec:discussion}

\subsection{Network Architecture Considerations}
While the results for the direct ConvNet architecture were marginally better than the kinematic ConvNet with variable lengths, the latter has other advantages. First, there can be value in getting a skeletal model from the network, providing a more complete set of parameters including 20 angular DOFs and a total of 17 joint positions. Second, unsurprisingly, requiring that outputs from the ConvNet satisfy kinematic constraints means that outputs will be constrained to plausible looking body poses. Without those constraints the ConvNet could produce unrealistic outputs.

While limiting our skeleton model to 20 angular DOFs promotes simplicity, it has some hindrance to generalizability. For example, the mid-spine joint lacks rotational DOFs about the y- and z-axes. Adding these DOFs would allow the model to account for rolling to a different posture and laying sideways in bed.

\subsection{Data Augmentation Challenges}
Data augmentation cannot easily account for a person sliding up and down in a bed that is not flat. Vertical shifting augmentation for non-flat beds would not match the physical effects of shifting a person on the pressure mat. Augmentation by scaling also has problematic implications, because a much smaller or larger person may have a weight distribution that would not scale linearly at the bending point of the bed. Simulation might resolve these issues by simulating placing a weighted human model of variable shape and size placed anywhere on a simulated pressure mat, with many possibilities of bed configurations.

\subsection{Dataset Considerations}
The posture and range of paths traversed by participants may not be representative of other common poses, and we expect our method to have limited success in generalizing to body poses not seen or rarely seen in the dataset. While a participant is moving their arm across a specified path, other joints remain nearly static, which over-represents poses with the arms adjacent to the chest and legs straight. We found over-represented poses to generally have a lower uncertainty. Interestingly, with our current sampling and training strategy, over-representation and under-representation is based on the percentage of images in the dataset a joint or set of joints is in a particular configuration. Additional epochs of training or directly scaling the size of the dataset does not change these effects on uncertainty of pose representation. Our method could be improved by using weighting factors or sampling strategies to compensate for this effect. 

\begin{figure}
\centering
\includegraphics[width=\columnwidth]{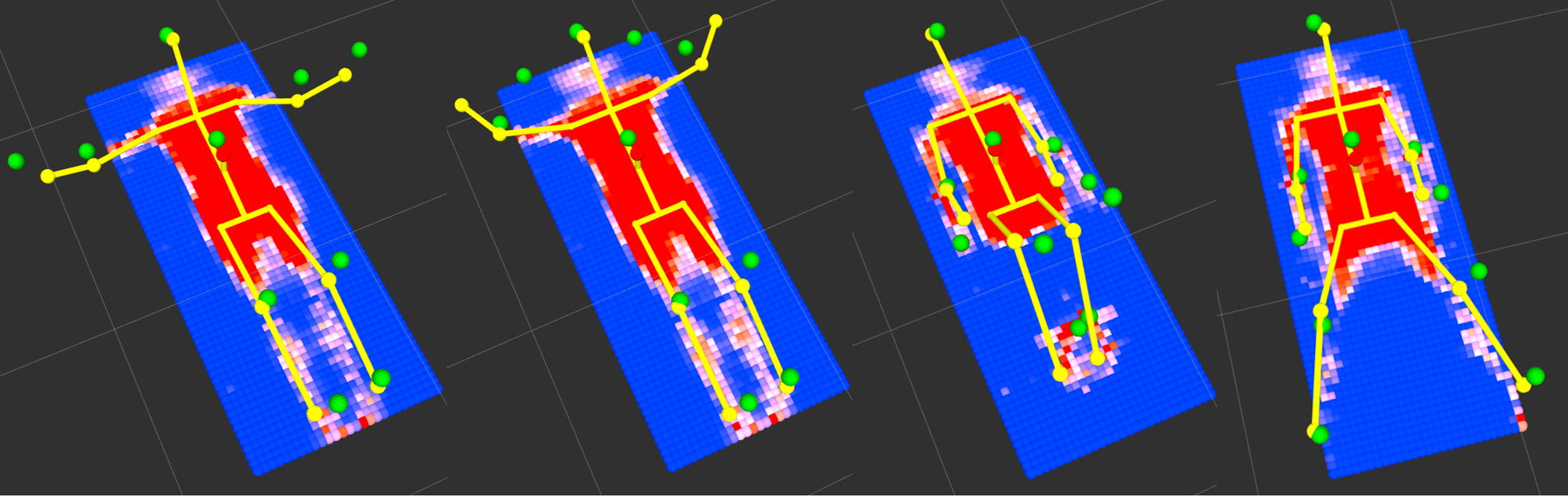}
\caption{Dual arm and dual leg traversals. While these are excluded from the training set, our methods can provide a reasonable pose estimate.}
\label{fig:dual_traversal}
\vspace{-3mm}
\end{figure}

In our evaluation, some limb poses occur in separate training images, but do not occur in the same training image. For example, we recorded one participant moving both arms and both legs simultaneously. Fig. \ref{fig:dual_traversal} shows that our method has some ability to estimate these poses.

The skeleton model has offset error in addition to the ground truth error reported. While we attempted to compensate for the chest marker offset, the other markers were more challenging. This may have caused some inaccuracy in the link length approximations.

\subsection{Removal of High Variance Joints}

\begin{figure}
\centering

\includegraphics[width=\columnwidth]{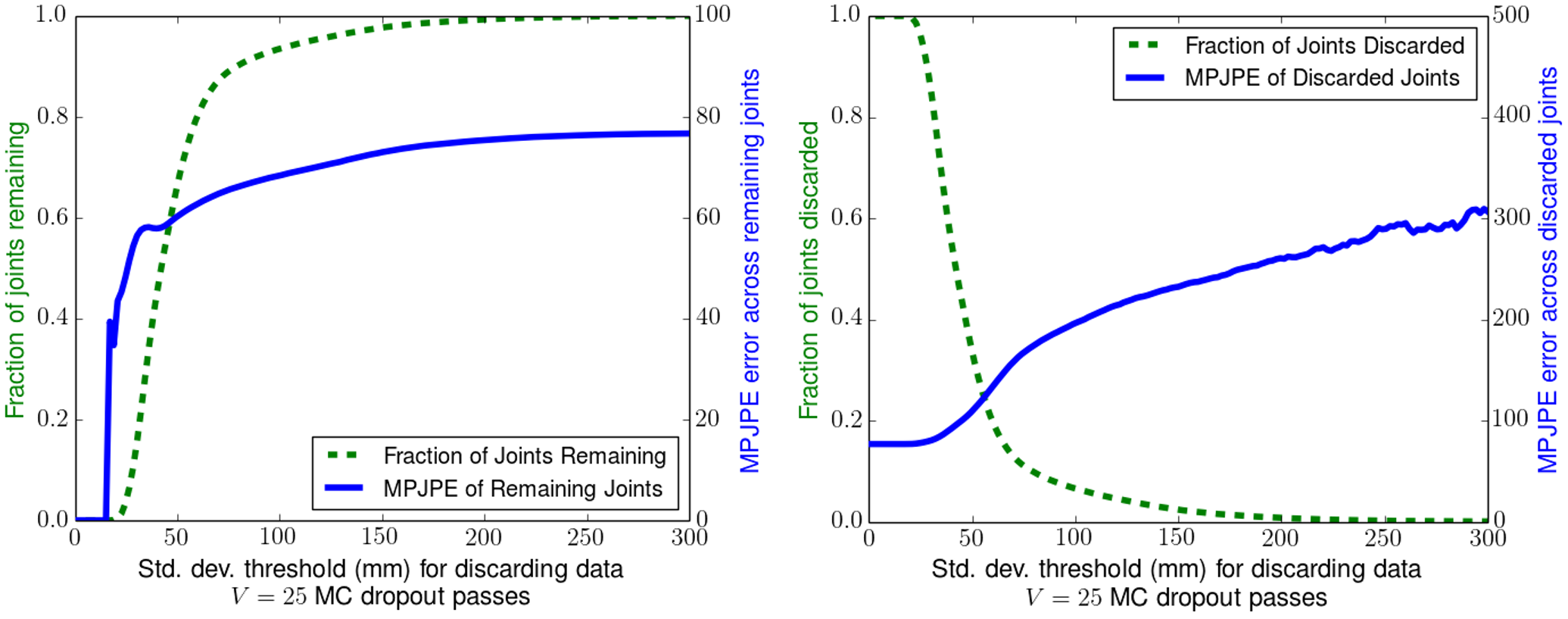}
\caption{Discarding joints with a higher uncertainty can decrease error, with a tradeoff to the number of joints remaining.}
\label{fig:out}
\end{figure}

Fig.~\ref{fig:out} shows that joints with high uncertainty have a higher average error. Discarding these joint estimates can decrease the average error of the model. As an example application, a robot using our method's estimated poses for task and motion planning might want to require low uncertainty before plan execution.

\section{Demonstration with PR2 Robot}
We conducted a demonstration of how our method could inform an assistive robot trying to reach part a person's body. We conducted this study with approval from the Georgia Institute of Technology Institutional Review Board (IRB), and obtained informed consent from an participant. 
We recruited a single able-bodied participant who used a laptop computer running a web interface from \cite{grice2016assistive} to command a PR2 robot to move its end effector to their left knee and to their left shoulder. The robot's goal was based on the estimated pose of the person's body from our ConvNet with kinematic model regressing to link lengths. For the knee position, the participant raised her knee to the configuration shown in the 2nd image of Fig. \ref{fig:training_poses_supine}~(f), and the participant was in the seated posture for both tasks. Using our 3D pose estimation method, the robot was able to autonomously reach near both locations. Fig~\ref{fig:demo_shoulder} shows the robot reaching a shoulder goal while the participant is occluded by bedding and an over-bed table.

\section{Conclusion}
In this work, we have shown that a pressure sensing mat can be used to estimate the 3D pose of a human in different postures of a configurable bed.  We explored two ConvNet architectures and found that both significantly outperformed data-driven baseline algorithms. Our kinematically embedded ConvNet with link length regression provided a more complete representation of a 17-joint skeleton model, adhered to anthropomorphic constraints, and was able to adjust to participants of varying anatomy.  We provided an example where joints on limbs raised from the pressure mat had a higher uncertainty than those in contact. We demonstrated our work using a PR2 robot.

\section*{Acknowledgment}

\small
\textit{We thank Wenhao Yu for his suggestions to this work. This material is based upon work supported by the National Science Foundation Graduate Research Fellowship Program under Grant No. DGE-1148903, NSF award IIS-1514258, NSF award DGE-1545287, AWS Cloud Credits for Research, and the National Institute on Disability, Independent Living, and Rehabilitation Research (NIDILRR), grant 90RE5016-01-00 via RERC TechSAge. Any opinion, findings, and conclusions or recommendations expressed in this material are those of the author(s) and do not necessarily reflect the views of the National Science Foundation. Dr. Kemp is a cofounder, a board member, an equity holder, and the CTO of Hello Robot, Inc., which is developing products related to this research. This research could affect his personal financial status. The terms of this arrangement have been reviewed and approved by Georgia Tech in accordance with its conflict of interest policies.}

\bibliographystyle{IEEEtran}
\bibliography{main}

\end{document}